\documentclass[letterpaper,10pt, conference]{ieeeconf}

\usepackage{setspace}
\usepackage{url}
\usepackage[utf8]{inputenc}
\usepackage[T1]{fontenc}
\usepackage{times}
\usepackage{graphicx}		% For \includegraphics
\usepackage{wrapfig}
\usepackage[format=plain,font=footnotesize,labelfont=bf,labelsep=period]{caption}
\usepackage{sidecap} % For sidecaptions for figures
\usepackage[export]{adjustbox}
\usepackage{subcaption}
\usepackage[font=small]{caption}
\usepackage{float}
\usepackage{comment}

\usepackage{amsmath} % assumes amsmath package installed
\usepackage{amssymb}  % assumes amsmath package installed
\usepackage{amsthm}
\usepackage{mathtools}

\usepackage{xcolor}

\usepackage[normalem]{ulem}
\usepackage{paralist}	% For enumerations with better titled numbers
\usepackage[space]{grffile} % Space in filenames
\usepackage{color}
\let\labelindent\relax
\usepackage{enumitem}
\usepackage{bm}
\usepackage{cancel}
\usepackage{hhline}
\usepackage[c2 , nocomma]{optidef}
\usepackage{oubraces}
\usepackage{textcomp}
\usepackage{booktabs} % For \toprule, \midrule, \bottomrule
\usepackage{longtable}

\usepackage{moreverb, url}

\usepackage[ruled,vlined]{algorithm2e}
\usepackage{stackengine}
\usepackage{dirtytalk}

\usepackage{array}
\usepackage{verbatim}
\usepackage{upgreek}

\usepackage{amsfonts}
\usepackage{lipsum}
\usepackage{cite}
\usepackage{hyperref}

\IEEEoverridecommandlockouts
\begin{document}
\spacing{1}
\title{ \bf Real-Time Disturbance-Aware Koopman Model Predictive Control for Quadrotors in Spatially Varying Winds}
\vspace{2mm}
\author{%
Stella Kombo$^{1}$, Masih Haseli$^{1}$, Alejandro A. Stefan-Zavala$^{1}$, Peter I. Renn$^{1}$, Morteza Gharib$^{1}$, and Joel W. Burdick$^{1}$%
%\thanks{*This work was supported by the Defense Advanced Research Projects Agency (DARPA) under the LINC program.}%
\thanks{$^{1}$S. Kombo, M. Haseli, A. A. Stefan-Zavala, P. I. Renn, M. Gharib, and  J. W. Burdick are with the Division of Engineering and Applied Science, California Institute of Technology, Pasadena, CA 91125, USA.
${\tt\small \{skombo, mhaseli, aastefan, prenn, mgharib, jburdick\}@caltech.edu}$}%
}

\maketitle
\begin{abstract}
Reliable quadrotor flight in disturbed environments requires predictive models that remain accurate as unmodeled aerodynamic effects change during deployment. We introduce a disturbance-aware Koopman Control Family (DA-KCF) framework that preserves an offline-identified nominal model while learning only 
a disturbance-induced correction online. A reference-dependent KCF is first learned from simulation data and calibrated using limited hardware data. During deployment, the nominal KCF remains fixed while a lightweight residual ensemble learns a low-dimensional disturbance correction from causal state--command--state transitions. The Adaptive DA-KCF model is incorporated into MPC, where disturbance corrections are evaluated along the prediction horizon to optimize position and yaw references sent to the onboard stabilizing controller. Hardware experiments on a nano-quadrotor under uniform wind and a spatially varying wake demonstrate that online adaptation improves multi-step prediction and trajectory tracking as disturbance-induced model mismatch increases, while satisfying real-time control requirements.
\end{abstract}
%----------------------------------------%
\section{Introduction}\label{intro}
Autonomous aerial robots are increasingly being considered for inspection, delivery, monitoring and human mobility tasks in densely built environments~\cite{rob}. Such missions require multi-rotors to operate around buildings, above rooftops, through corridors and other confined structures while maintaining accurate trajectory tracking and satisfying state and actuation constraints. Reliable flight in these settings remains challenging since quadrotors are nonlinear, underactuated systems whose performance can degrade when aerodynamic effects are not adequately represented by the prediction or control model~\cite{neurofly2024,erman2018}. The challenge is particularly pronounced in urban environments, where interactions between incoming flow and surrounding structures generate spatially varying wakes, vortices, recirculation zones and accelerated flows around building edges~\cite{wind_char}. Consequently, the disturbances experienced by the aerial vehicle can vary significantly along a single flight trajectory. 

Aerodynamic disturbances have previously been addressed at several levels of aerial-robot modeling and control. Classical approaches mainly rely on feedback to reject external disturbances and modeling errors during flight~\cite{mellinger2011minimumsnap}. Analytic aerodynamic models, such as linear/quadratic rotor-drag terms, augment nominal dynamics with low-order aerodynamic effects while remaining compatible with real-time control~\cite{Faessler2018}. Data-driven approaches instead learn residual dynamics or full-state transition models from flight data, capturing complex effects that are difficult to model analytically. These learned effects can improve prediction and control for high-performance drone flight~\cite{bauersfeld2021neurobem, torente, chee2022knodempc}. Although such models can be identified offline and used efficiently online, their accuracy degrades when the vehicle encounters dynamics not represented in the identification data. 

A complementary line of work uses measured or predicted spatial flow information to aid wind-aware trajectory planning and penalize paths that enter adverse flow regions~\cite{ware2016windfield}. High-fidelity computational fluid dynamics can resolve complex urban flow structures~\cite{galway2012urban}, but their computational costs limit direct use in online prediction and high-rate control~\cite{galway2011urbanwind}. Learned flow surrogates provide a tractable alternative for wind-informed planning~\cite{stefanzavala2026fanarray, renn2026windinformed}. However, disturbed regions cannot always be avoided when geometric constraints, mission requirements or unknown wakes force the vehicle to traverse them. When this occurs, unmodeled disturbances can induce persistent mismatch between nominal predictions and the realized vehicle response, degrading tracking and stability. Adaptive, observer-based and learning-based controllers compensate for these effects by estimating disturbances, adapting model parameters or learning residual corrections online~\cite{drewadaptivecontrol,benevides2022disturbance,neurofly2024}. For predictive control, these corrections must also be propagated over the prediction horizon without requiring online retraining of the full dynamics model. This motivates learning a nominal model offline from broad system excitation, while adapting to unmodeled disturbance-induced discrepancies during deployment.   %------------------------------------%
\begin{figure}[t]
    \centering
    \includegraphics[width=0.60\linewidth]{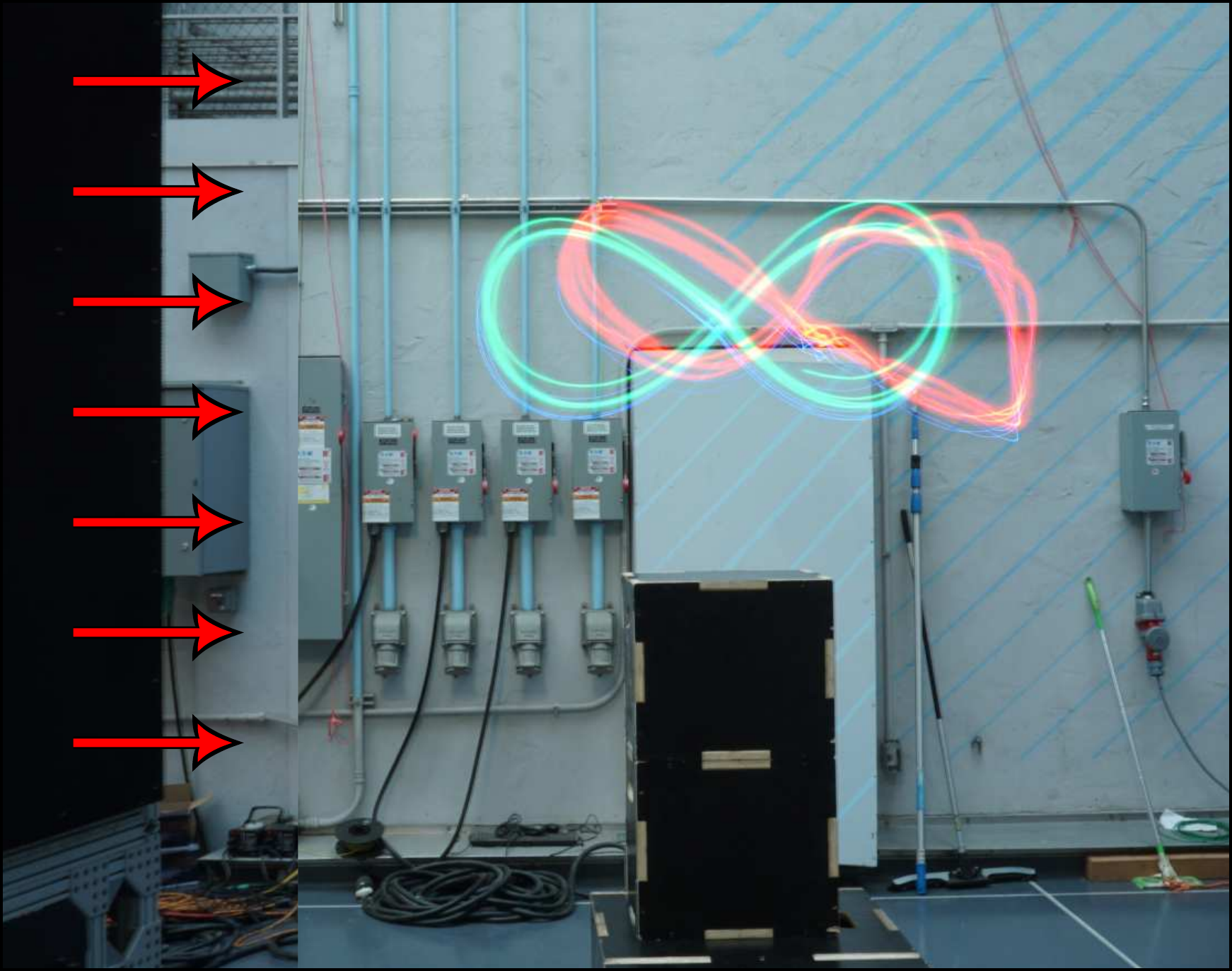}
    \captionsetup{font=footnotesize}\caption{Drone figure-eight trajectory tracking under aerodynamic disturbance. The fan array seen on the left side of the photo generates a predominantly unidirectional flow (red arrows). The overlaid trajectories illustrate the adaptive (green) and non-adaptive (red) controller responses.}
    \label{fig:overview}
    \vspace{-6.5mm}
\end{figure}
%------------------------------------%

Koopman-based models provide a structured nominal predictor that can be learned offline and used efficiently online, making them potentially suitable for disturbance augmentation. By representing nonlinear dynamics through observables in a function space, finite-dimensional approximations can be incorporated into model predictive control~\cite{williams2015edmd, korda2018koopman} and have been applied to quadrotor prediction and trajectory tracking~\cite{folkestad2022koopnet, rajkumar2025realtime}. The Koopman Control Family (KCF)~\cite{haseli2026koopmancontrol} represents nonlinear and state-dependent control systems via a family of linear operators parameterized by the control input. However, existing KCF methods do not explicitly represent disturbances that arise during deployment~\cite{haseli2025extensions}. 

Koopman methods have also been extended to disturbance estimation and online adaptation. Disturbance-observer approaches learn disturbance models for online compensation~\cite{jia2024evolver}, while adaptive Koopman methods update residual or lifted dynamics~\cite{singh2025adaptive,zhou2025noregret}. For quadrotors, related work either switches among pre-identified Koopman models across operating conditions~\cite{oh2024koopmanwind} or augments nominal predictors with residual corrections~\cite{todde2026residual}. These approaches do not formulate online adaptation as a disturbance-dependent extension of an input-dependent KCF. This leaves open how to preserve a pretrained nominal KCF while adapting only to disturbance-induced discrepancies online within real-time predictive control, without retraining the full model. 
%----------------------------------%
\subsubsection*{\bf{Contribution}}
To address this gap, we introduce a {\em disturbance-aware Koopman Control Family} (DA-KCF) and apply it to real-time predictive control of multi-rotors. We first identify a nominal KCF from simulation data and calibrate it using hardware data, capturing the coupled vehicle-onboard controller response while keeping the nominal model fixed during deployment. We then adopt a structured separable approximation $A(u,d)\approx A_{\mathrm{c}}(u)+A_d(d)$, where $A_{\mathrm{c}}(u)$ is the nominal reference-dependent KCF and $A_d(d)$ is a disturbance-dependent modification. For real-time deployment, the disturbance-dependent effect is realized through a low-dimensional correction learned online and propagated through the predictive horizon, while the calibrated nominal KCF and onboard stabilization controller remain fixed. We validate the method on a nano-quadrotor under previously unmodeled aerodynamic disturbances. Experiments include figure-eight tracking in uniform fan-array flows and traversal through a spatially varying wake generated by a building-like obstacle. The results demonstrate online adaptation of the predictive dynamics, improved trajectory tracking as disturbance-induced model mismatch increases, and real-time execution within the control deadline.
\subsubsection*{Notation}
$\mathbb N$, $\mathbb R$, and $\mathbb C$ denote the natural, real, and complex numbers, respectively. We use $\|\cdot\|_2$ and $\|\cdot\|_F$ for the Euclidean and Frobenius norms. $I_n$ denotes the $n\times n$ identity matrix. For vectors $v_1,\ldots,v_q$, $\operatorname{col}(v_1,\ldots,v_q):=\begin{bmatrix}v_1^\top & \cdots & v_q^\top\end{bmatrix}^{\top}$. For a matrix $M$, $\operatorname{vec}(M)$ denotes the vector obtained by stacking the entries of $M$ row-wise. For square matrices $M_1, \ldots,M_r$, $\operatorname{blkdiag}(M_1, \ldots,M_r)$ denotes the block-diagonal matrix with blocks $M_1, \ldots,M_r$.
For a finite set $\mathcal B$, $|\mathcal B|$ denotes the number of its elements. For a matrix $Q\succeq0$, $\|v\|_Q:=\sqrt{v^\top Qv}$. For functions $f$ and $g$ with compatible domains and co-domains, $f\circ g$ denotes their composition. For $H=(h_1,\ldots,h_{n_H})^\top$, $\operatorname{span}(H)$ denotes the vector space spanned by its components. $x^{+}$ denotes the successor state of $x$ under a discrete-time dynamical system.
% $\mathbb{N}$, $\mathbb{N}_0$, $\mathbb{R}$, and $\mathbb{C}$ denote the sets of natural, nonnegative integer, real, and complex numbers, respectively. For a vector $v$ and matrix $M$, $\|v\|_2$ and $\|M\|_F$ denote the respective Euclidean and Frobenius norms, and $I_n$ is the $n\times n$ identity matrix. For vectors $v_1,\ldots,v_q$, $\operatorname{col}(v_1,\ldots,v_q):=\begin{bmatrix}v_1^\top & \cdots & v_q^\top\end{bmatrix}^{\top}$. For a matrix $M$, $\operatorname{vec}(M)$ denotes the vector obtained by stacking the entries of $M$ row-wise. For square matrices $M_1, \ldots,M_r$, $\operatorname{blkdiag}(M_1, \ldots,M_r)$ denotes the block-diagonal matrix with blocks $M_1, \ldots,M_r$. For a finite set $\mathcal{B}$, $|\mathcal{B}|$ denotes the number of elements in $\mathcal{B}$ (its cardinality). For $Q\succeq0$, define the weighted quadratic semi-norm $\|v\|_Q:=\sqrt{v^\top Qv}$. For functions $f$ and $g$ with compatible domains and co-domains, $f\circ g$ denotes their composition. For $H=(h_1,\ldots,h_{n_H})^\top$, $\operatorname{span}(H)$ denotes the vector space spanned by its components. $x^{+}$ denotes the successor state of $x$ under a discrete-time dynamical system.
%----------------------------------------%
\section{Preliminaries}\label{sec:preliminaries}
This section briefly reviews the Koopman Operator for dynamical systems and the Koopman Control Family (KCF) framework used throughout the paper.
%----------------------------------------%
\vspace{-2mm}
\subsection{Koopman Operator}
Consider the discrete-time nonlinear dynamical system
\begin{equation}
    x^{+}=T(x), \qquad x\in\mathcal{X}\subseteq\mathbb{R}^{n},
    \label{eq:autonomous_sys}
\end{equation}
where $T:\mathcal{X}\rightarrow\mathcal{X}$ is the state-transition map. Let $\mathcal{F}$ be a vector space over $\mathbb{C}$ of {\em observables} $f:\mathcal{X}\rightarrow\mathbb{C}$ that is closed under composition with $T$. The Koopman operator $\mathcal{K}:\mathcal{F}\rightarrow\mathcal{F}$ is defined by $\mathcal{K}f:=f\circ T$, such that $(\mathcal{K}f)(x)=f(x^{+})$. Although $T$ may be nonlinear, $\mathcal{K}$ is a linear operator on space $\mathcal{F}$~\cite{koopman1931hamiltonian,mezic2005spectral}. Let $H:\mathcal{X}\rightarrow\mathbb{C}^{n_H}$ be a vector-valued map whose elements span a finite-dimensional subspace $\mathcal{S}=\operatorname{span}(H)\subseteq\mathcal{F}$. If $\mathcal{S}$ is invariant under $\mathcal{K}$, there exists $A\in\mathbb{C}^{n_H\times n_H}$ such that, for all $x\in\mathcal{X}$, \begin{equation}
H(x^{+})=H(T(x))=(H\circ T)(x)=AH(x).
\label{eq:koopman_lifted}
\end{equation}
Defining $z:=H(x)$, the dynamics evolve linearly in the lifted coordinates as $z^{+}=Az$.\footnote{A finite-dimensional invariant subspace does not need to exist. One seeks an approximation $H(T(x))\approx AH(x)$, for example using data-driven methods such as Extended Dynamic Mode Decomposition (EDMD)~\cite{williams2015edmd}.}
%----------------------------------------%
\subsection{Koopman Control Family}
We briefly review the Koopman Control Family (KCF) framework introduced in~\cite{haseli2026koopmancontrol}. Consider the discrete-time nonlinear control system $x^{+}=F(x,u)$, where $x\in\mathcal{X}\subseteq\mathbb{R}^{n_x}$, $u\in\mathcal{U}\subseteq\mathbb{R}^{n_u}$, and $F:\mathcal{X}\times\mathcal{U}\rightarrow\mathcal{X}$. For each constant input $u^\ast\in\mathcal{U}$, define $F_{u^\ast}:\mathcal{X}\rightarrow\mathcal{X}$ by $F_{u^\ast}(x):=F(x,u^\ast)$. If $\mathcal{F}$ is closed under composition with every $F_{u^\ast}$, the associated KCF is the family of Koopman operators $\{\mathcal{K}_{u^\ast}:\mathcal{F}\rightarrow\mathcal{F}\}_{u^\ast\in\mathcal{U}}$, where $\mathcal{K}_{u^\ast}f:=f\circ F_{u^\ast}$. Let $H:\mathcal{X}\rightarrow\mathbb{C}^{n_H}$ be a vector-valued map whose elements span a finite-dimensional common invariant subspace of the KCF. Then there exists a matrix-valued map $A:\mathcal{U}\rightarrow\mathbb{C}^{n_H\times n_H}$ such that, for all $x\in\mathcal{X}$ and $u\in\mathcal{U}$,
\begin{equation}
   H(x^+) = H(F(x,u))=A(u)H(x).
    \label{eq:kcf_input_state}
\end{equation}
Thus, the lifted dynamics are linear for fixed $u$, while $A(u)$ may depend nonlinearly on the input\footnote{The input-state separable representation includes lifted linear, bilinear, and switched-linear Koopman models as special cases~\cite{haseli2026koopmancontrol}.}. When no informative finite-dimensional common invariant subspace exists, one instead seeks approximations of the form~\eqref{eq:kcf_input_state}; the optimal approximations and sharp accuracy bounds on the function prediction error can be found in~\cite[Sec.~8]{haseli2026koopmancontrol} and~\cite{haseli2026control}.
%----------------------------------------%
\section{Disturbance-Aware Koopman Control Family (DA-KCF)}
\label{sec:disturbance-kcf}
The KCF framework naturally accommodates multiple input types. Let the composite input be $\bar u:=\operatorname{col}(u_{\mathrm{c}},d)$, where $u_{\mathrm{c}}\in\mathcal{U}_{\mathrm{c}}$ denotes the control and $d\in\mathcal{D}\subseteq\mathbb{R}^{n_d}$ denotes a disturbance. Consider the control system $x^{+}=F(x,u_{\mathrm{c}},d)$. For each fixed pair $(u_{\mathrm{c}}^\ast,d^\ast)\in\mathcal{U}_{\mathrm{c}}\times\mathcal{D}$, define $F_{u_{\mathrm{c}}^\ast,d^\ast}(x):=
F(x,u_{\mathrm{c}}^\ast,d^\ast)$. If $\operatorname{span}(H)$ is a finite-dimensional common invariant subspace of the composite-input KCF, there exists a matrix-valued map $A:\mathcal{U}_{\mathrm{c}}\times\mathcal{D}\rightarrow\mathbb{C}^{n_H\times n_H}$ such that
%\footnote{When no informative finite-dimensional common invariant subspace exists, one may instead seek a finite-dimensional approximation of the KCF relation and characterize its accuracy through subspace invariance proximity~\cite{haseli2026koopmancontrol}.}
\begin{equation}
H(x^{+})=H(F(x,u_{\mathrm{c}},d))=
A(u_{\mathrm{c}},d)H(x).
\label{eq:disturbance_kcf}
\end{equation}
The lifted dynamics remain linear for each realized pair $(u_{\mathrm{c}},d)$, while the evolution may depend nonlinearly on both input components. To separate the nominal control-dependent component $A_{\mathrm{c}}(u_{\mathrm{c}})$ from the disturbance-dependent effect $A_d(d)$, we consider the approximation 
\begin{equation}
A(u_{\mathrm{c}},d)\approx A_{\mathrm{c}}(u_{\mathrm{c}})+A_d(d).
\label{eq:additive_disturbance_kcf}
\end{equation}
Ideally, if the disturbance parameter $d$ is directly measurable, $A_d(d)$ could in principle be identified offline from experiments or simulations spanning a range of disturbance conditions. 
However, in the setting considered here, $d$ is not directly observed. Instead, we realize its effect through a correction learned from observed model mismatch\footnote{For the quadrotor model considered below, $u_{\mathrm{c}}$ corresponds to the commanded-reference input $u$: we hereafter omit the subscript $\mathrm{c}$.}.
%----------------------------------------%
\section{System Model and Problem Formulation} 
\label{sec:problem}
%----------------------------------------%
\subsection{Quadrotor and Internal-Controller Dynamics}
\label{subsec:quad_system}
Many quadrotors, such as the Crazyflie considered here, use an onboard low-level controller to stabilize the vehicle from commanded position and yaw references. We therefore model the quadrotor and its onboard controller as a discrete-time control system, with these commanded references acting as the control $u_{\mathrm{c}}$ in Eq.~\eqref{eq:disturbance_kcf}. To define the system state, let the rigid-body and auxiliary states be $x_k^{\mathrm{rb}}:=\operatorname{col}\!\left(p_k,v_k,\operatorname{vec}(R_k),\omega_k\right)$ and $x_k^{\mathrm{aux}}:=\operatorname{col}\!\left(\Omega_k,V_{b,k},\chi_k,c_k,\rho_k\right)$, respectively. 
The augmented state and reference inputs are denoted by
\begin{subequations}
\begin{align}
x_k &:= \operatorname{col}\!\left(x_k^{\mathrm{rb}},x_k^{\mathrm{aux}}\right)
\in \mathcal{X}\subseteq\mathbb{R}^{n_x},
\label{eq:quad_state}\\
u_k &:= \operatorname{col}\!\left(p_k^{\mathrm{cmd}},\psi_k^{\mathrm{cmd}}\right)
\in \mathcal{U}\subseteq\mathbb{R}^{n_u}.
\label{eq:quad_input}
\end{align}
\end{subequations}
Here, $p_k,v_k,\omega_k\in\mathbb{R}^{3}$ denote inertial-frame position, translational velocity, and body angular velocity; $R_k\in\mathrm{SO}(3)$ is the body-to-inertial rotation; and $\Omega_k\in\mathbb{R}^{4}$ and $V_{b,k}\in\mathbb{R}$ denote motor speeds and battery voltage, respectively. The variables $\chi_k\in\mathbb{R}^{40}$, $c_k\in\mathbb{R}^{9}$, and $\rho_k\in\mathbb{R}^{4}$ collect internal-controller states and signals, command-target states, and motor PWM signals, respectively. In our implementation, $n_x=76$ and $n_u=4$. The augmented state includes the internal controller memory required for a Markovian reference-to-state representation. Consistent with Sec.~\ref{sec:disturbance-kcf}, let $F_0(x,u):=F(x,u,0)$ denote the nominal dynamics:
\begin{equation}
x_{k+1}=F_0(x_k,u_k).
\label{eq:nominal_closed_loop}
\end{equation}
During deployment, unmodeled dynamics and aerodynamic disturbances may alter this evolution:
\begin{equation}
x_{k+1}
=
F_0(x_k,u_k)+\Delta F_k,
\label{eq:disturbed_closed_loop}
\end{equation}
where $\Delta F_k\in\mathbb{R}^{n_x}$ denotes the unknown change in the
state transition relative to the nominal dynamics. In the lifted representation of Eq.~\eqref{eq:additive_disturbance_kcf}, we seek to capture this unknown effect through the disturbance-dependent KCF term $A_d(d_k)$.
%----------------------------------------%
\subsection{Problem Statement}
\label{subsec:prob_state}
Quadrotors operating in urban environments may encounter spatially varying aerodynamic disturbances. We seek a predictive KCF model of the form in Eq.~\eqref{eq:disturbance_kcf}, approximated by the additive structure in Eq.~\eqref{eq:additive_disturbance_kcf}. However, during deployment, the disturbance $d_k$ is not directly measurable\footnote{Disturbance $d$ need not have zero-mean stochastic behavior; it may represent persistent or structured effects such as aerodynamic wind disturbances.}.

We must therefore infer the disturbance-induced modification from the transition history $\mathcal{H}_k:=\{(x_i,u_i,x_{i+1})\}_{i=0}^{k-1}$. This modification must be learned \emph{online} from the observed vehicle response and remain computationally tractable for real-time use within the MPC.
%----------------------------------%
\begin{figure*}
    \centering
    \includegraphics[width=0.75\linewidth]{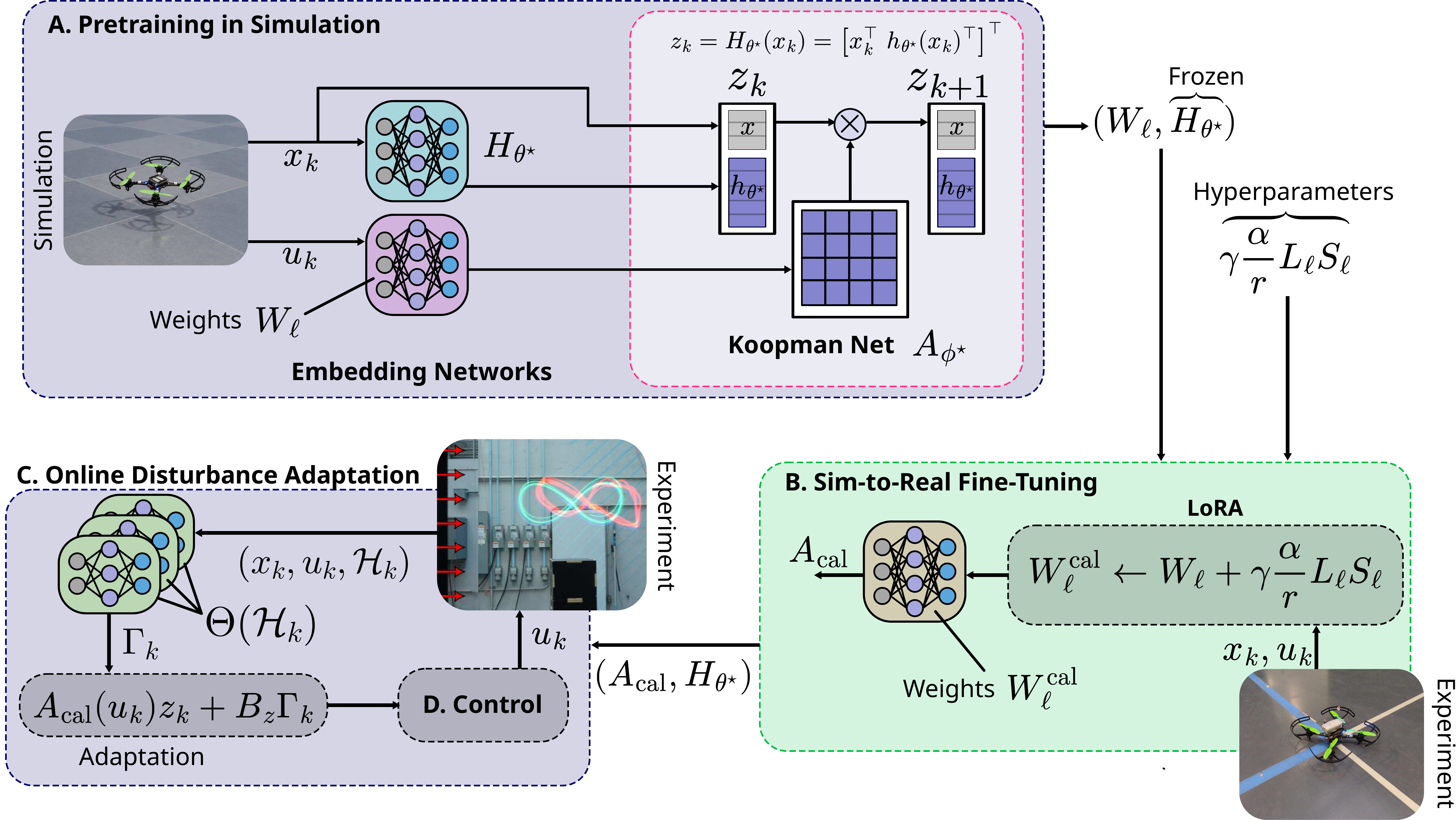}
    \captionsetup{font=footnotesize}
    \caption{Overview of the proposed disturbance-aware KCF pipeline: nominal KCF identification in simulation, sim-to-real fine-tuning and online disturbance adaptation for predictive control.}
    \label{fig:hero}
    \vspace{-5mm}
\end{figure*}
%----------------------------------%
\section{Methodology}
\label{sec:Methodology}
The proposed pipeline is summarized in Fig.~\ref{fig:hero}.
\textbf{(A)} We first identify the nominal KCF using diverse simulation data, reducing the need for extensive hardware data collection.
\textbf{(B)} We then calibrate the nominal KCF offline using Low-Rank Adaptation (LoRA) on a smaller hardware dataset to reduce the sim-to-real mismatch while preserving the learned lifting map.
\textbf{(C)} During deployment, we provide an additive online adaptation to the calibrated KCF model to account for disturbance.
\textbf{(D)} The resulting Adaptive DA-KCF model is used within the MPC for real-time predictive control.
%----------------------------------%
\subsection{Pretraining in Simulation}
\label{subsec:nominal-kcf}
We identify the nominal reference-to-state predictor by learning a finite-dimensional approximation of the KCF relation in Eq.~\eqref{eq:kcf_input_state}, with $u_k$ denoting the commanded-reference input. Accurate Koopman identification requires sufficiently informative state-reference data~\cite{williams2015edmd}, which is costly to collect exclusively on hardware for the large augmented state considered here. We therefore pretrain the nominal KCF in simulation. We use a MuJoCo/MJX model based on the Crazyflie Brushless platform~\cite{graefe2026crazyflie,crazyfliebrushjax}, augmented with the cascaded position, velocity, attitude, and attitude-rate controllers of the Bitcraze Crazyflie firmware~\cite{bitcraze_firmware}. The simulation dataset contains $250{,}000$ transitions ($83.3$~min) sampled at $50$~Hz, $\mathcal{D}_{\mathrm{sim}}:=\{(x_k^{(q)},u_k^{(q)},x_{k+1}^{(q)})\}_{q,k}$, where $q$ indexes trajectories and $k$ indexes time within each trajectory.

We parameterize the lifting map $H$ in Eq.~\eqref{eq:kcf_input_state} by $H_\theta$ and the matrix-valued function $A(u)$ by $A_\phi(u)$, and learn them jointly from $\mathcal{D}_{\mathrm{sim}}$ such that $H_\theta(x_{k+1})\approx A_\phi(u_k)H_\theta(x_k)$. The state-inclusive lift is
\begin{equation}
z_k=H_\theta(x_k):=
\begin{bmatrix}
x_k\\
h_\theta(x_k)
\end{bmatrix}
\in\mathbb{R}^{n_H},
\label{eq:nominal_lift}
\end{equation}
where $h_\theta:\mathbb{R}^{n_x}\rightarrow\mathbb{R}^{n_H-n_x}$ contains the learned observables. We augment the $76$ state coordinates with $16$ learned observables, selected empirically. Additional observables beyond $16$ provided no clear prediction improvement while increasing operator size. Thus, $n_H=92$ and $A_\phi(u)\in\mathbb{R}^{92\times92}$. 

For clarity, we omit the trajectory index $q$ below. Starting from measured state $x_k$, the model predicts $h_p$ steps without replacing intermediate predictions with measured states:
\begin{equation}
\begin{aligned}
\widehat z_k
&=H_\theta(x_k),\\
\widehat z_{k+j}
&=A_\phi(u_{k+j-1})\widehat z_{k+j-1},
\quad j=1,\ldots,h_p.
\end{aligned}
\label{eq:kcf_rollout}
\end{equation}
For a batch $\mathcal{B}$ of trajectory starting indices, we minimize
\begin{equation}
\begin{aligned}
\mathcal{L}_{\mathrm{KCF}}
=
&\frac{1}{|\mathcal{B}|}
\sum_{k\in\mathcal{B}}
\Bigg[
\frac{1}{n_H}
\left\|
\widehat z_{k+1}
-
H_\theta(x_{k+1})
\right\|_2^2\\
&+
\frac{\lambda_{\mathrm{ms}}}{h_p-1}
\sum_{j=2}^{h_p}
\frac{1}{n_H}
\left\|
\widehat z_{k+j}
-
H_\theta(x_{k+j})
\right\|_2^2
\Bigg].
\end{aligned}
\label{eq:kcf_loss}
\end{equation}
The first term penalizes one-step prediction error and the second penalizes error over the remaining horizon. The weight $\lambda_{\mathrm{ms}}\geq0$ sets the relative importance of the multi-step term. A curriculum strategy~\cite{bengio2009curriculum} progressively increases the prediction horizon to $h_p$ during training. The simulation-identified model, $z_{k+1}=A_{\phi^\star}(u_k)z_k$ with $z_k=H_{\theta^\star}(x_k)$, serves as the nominal KCF for hardware fine-tuning.
%-------------------------------------%
\subsection{Sim-to-Real Fine-Tuning}
\label{subsec:sim2real}
The simulation-based KCF is calibrated using hardware data $\mathcal{D}_{\mathrm{hw}}:=\{(x_k^{(q)},u_k^{(q)},x_{k+1}^{(q)})\}_{q,k}$ containing $86{,}025$ transitions ($28.7$~min of flight time) sampled at $50$~Hz. Low-rank adaptation (LoRA)~\cite{hu2022iclr-lora,han2024peft} is applied to four layers of $A_{\phi^\star}$ while keeping the lifting map $H_{\theta^\star}$ fixed. For each adapted layer $\ell$, $W_\ell^{\mathrm{cal}}=W_\ell+\Delta W_\ell$, where $\Delta W_\ell=(\alpha/r)L_\ell S_\ell$, $L_\ell$ and $S_\ell$ are trainable rank-$r$ factors, and $\alpha=r$. Thus, only the low-rank updates are learned from hardware. The updates minimize prediction error in $x_k^{\mathrm{rb}}$ at selected prediction horizons while limiting changes to the simulation-based model:
\begin{equation}
\mathcal{L}_{\mathrm{cal}}=\mathcal{L}_{\mathrm{physical}}+
\lambda_{\mathrm{reg}}
\sum_{\ell=1}^{4}
\frac{\|\Delta W_\ell\|_F^2}
{\|W_\ell\|_F^2+\epsilon},
\label{eq:lora_loss}
\end{equation}
where $\lambda_{\mathrm{reg}}\geq0$ controls the relative change penalty and $\epsilon>0$ ensures numerical stability. After fine-tuning, the resulting operator $A_{\mathrm{cal}}$ gives the nominal predictor
\begin{equation}
z_{k+1}=A_{\mathrm{cal}}(u_k)z_k,\qquad
z_k=H_{\theta^\star}(x_k).
\label{eq:calibrated_kcf}
\end{equation}
The calibrated KCF is then frozen and used in the next step for online adaptation and MPC.
%----------------------------------------%
%\vspace{-3.5mm}
\subsection{Online Disturbance Adaptation}
\label{subsec:adaptive-kcf}
 
The disturbance-dependent operator introduced in Eq.~\eqref{eq:additive_disturbance_kcf} acts on the full lifted dynamics of the closed-loop quadrotor system. Due to computational limits, we cannot adapt this full operator online. Instead, we restrict the adaptation to the drone's velocity coordinates\footnote{This is a structured modeling choice: aerodynamic forces directly affect the acceleration and therefore the velocity evolution. Furthermore, velocity is explicitly represented in the modeled state.}. To implement this restriction, we define the correction in the form $B_z \Gamma\!\left(x_k,u_k;\Theta(\mathcal H_k)\right)$, where
\begin{equation}
B_z:=
[0_{3\times3}\;\;I_3\;\;0_{3\times(n_H-6)}]^\top
\in\mathbb{R}^{n_H\times3},
\label{eq:residual_injection}
\end{equation}
and $\Gamma\!\left(x_k,u_k;\Theta(\mathcal H_k)\right)$ is the adaptation function. Here $x_k, u_k$ are the arguments of the adaptation function and $\Theta(\mathcal H_k)$ denotes the parameters updated online during adaptation\footnote{In implementation, the residual learner maintains a $250$-sample ($5$~s) FIFO buffer of past transitions for online adaptation, with $200$ samples used for training and the remaining $50$ for validation.}. We model the adaptation function, $\Gamma$, using an ensemble of $M=3$ residual networks, $g_{\Theta(\cdot)}$ as 
\begin{equation}
\begin{aligned}
\Gamma\!\left(x_k,u_k;\Theta(\mathcal H_k)\right)
\! = \!\frac{1}{M}\sum_{m=1}^{M}
g_{\Theta^{(m)}(\mathcal H_k)}
\!\left(\operatorname{col}(x_k,u_k)\right).
\end{aligned}
\label{eq:residual_estimator}
\end{equation}
\footnote{The residual network uses 23 selected physical-state features from $x_k$ and the $4$ commanded-reference inputs, yielding a $27$-dimensional input.}
The Adaptive DA-KCF predictor is therefore
\begin{equation}
z_{k+1}=A_{\mathrm{cal}}(u_k)z_k+B_z\Gamma_k,
\qquad
z_k=H_{\theta^\star}(x_k).
\label{eq:adaptive_kcf}
\end{equation}
To express the corrected predictor in Eq.~\eqref{eq:adaptive_kcf} in the matrix-valued DA-KCF form of Eq.~\eqref{eq:additive_disturbance_kcf}, introduce the homogeneous coordinate $\widetilde z_k:=\operatorname{col}(z_k,1)$. Then
\begin{equation}
\widetilde z_{k+1}
=
\left[
\widetilde A_{\mathrm{cal}}(u_k)
+
\widetilde A_{\Gamma}(\Gamma_k)
\right]\widetilde z_k,
\label{eq:adaptive_disturbance_kcf}
\end{equation}
where
\[
\widetilde A_{\mathrm{cal}}(u_k)=
\begin{bmatrix}
A_{\mathrm{cal}}(u_k) & 0\\
0^\top & 1
\end{bmatrix},
\qquad
\widetilde A_{\Gamma}(\Gamma_k)=
\begin{bmatrix}
0 & B_z\Gamma_k\\
0^\top & 0
\end{bmatrix}.
\]
Thus, Eq.~\eqref{eq:adaptive_disturbance_kcf} provides an online approximation of the DA-KCF in Eq.~\eqref{eq:disturbance_kcf} under the additive structure of Eq.~\eqref{eq:additive_disturbance_kcf}. The % calibrated 
nominal calibrated predictor % in Eq.
\eqref{eq:calibrated_kcf} remains fixed, while $\widetilde A_{\Gamma}(\Gamma_k)$  provides the history-adapted disturbance-dependent correction.
%---------------------------------------------%
\subsection{Control: Adaptive Model Predictive Control (MPC)}
\label{subsec:adaptive-mpc}

We embed the DA-KCF model from Eq.~\eqref{eq:adaptive_kcf} in a receding-horizon controller that optimizes the position and yaw references sent to the onboard controller. At time $k$, the measured state $x_k$ and the desired tracking reference $\{p^d_{i|k},v^d_{i|k},\psi^d_{i|k}\}_{i=1}^{h_p}$ over the prediction horizon are given. The optimizer selects the future commanded-reference sequence $U_k:=\{u_{0|k},\ldots,u_{h_p-1|k}\}$. For each predicted step, define the tracking error vector
\[
\epsilon_{i|k}
:=
\operatorname{col}\!\left(
\hat p_{i|k}-p^d_{i|k},
\hat v_{i|k}-v^d_{i|k},
e_{\psi,i|k}
\right),
\]
where $\hat p_{i|k}$, $\hat v_{i|k}$, and $\hat\psi_{i|k}$ are obtained from the predicted state and $e_{\psi,i|k}$ is the yaw error. Let $Q:=\operatorname{blkdiag}(Q_p,Q_v,q_\psi)\succeq0$ denote the tracking-weight matrix, with $Q_p,Q_v\in\mathbb{R}^{3\times3}$ and $q_\psi\geq0$. The MPC minimizes tracking error, mission reference, and command slew:
\begin{equation}
\begin{gathered}
\min_{U_k}
\sum_{i=0}^{h_p-1}
\left(
\|\epsilon_{i+1|k}\|_Q^2
+\|u_{i|k}-u^m_{i|k}\|_{R_u}^2
+\|\Delta u_{i|k}\|_{R_{\Delta u}}^2
\right)
\\[-1mm]
\begin{aligned}
\mathrm{s.t.}\quad
&z_{0|k}=H_{\theta^\star}(x_k),\\
&z_{i+1|k}
=A_{\mathrm{cal}}(u_{i|k})z_{i|k}
+B_z\widehat\Gamma_{i|k},
\quad i=0,\ldots,h_p-1,\\
&u_{i|k}\in\mathcal U,\qquad
\Delta u_{i|k}\in\Delta\mathcal U .
\end{aligned}
\end{gathered}
\label{eq:kcf_mpc}
\end{equation}
Here, $u^m_{i|k}$ is the mission reference and $\Delta u_{i|k}:=u_{i|k}-u_{i-1|k}$ with $u_{-1|k}:=u_{k-1}$. The matrices $R_u,R_{\Delta u}\succeq0$ weight reference deviation and slew, respectively. The residual-network parameters $\Theta(\mathcal H_k)$ remain fixed at each MPC solve. Using the learned residual map defined in Eq.~\eqref{eq:residual_estimator}, the correction at horizon step $i$ can be defined as:
\begin{equation}
\widehat\Gamma_{i|k}:=\Gamma\!\left(
\widehat x_{i|k},u_{i|k};
\Theta(\mathcal H_k)
\right),
\label{eq:mpc_residual}
\end{equation}
where $\widehat x_{i|k}$ is the predicted state at horizon step $i$. The disturbance correction is recomputed from the predicted state and commanded reference at each step in the MPC horizon. After solving Eq.~\eqref{eq:kcf_mpc}, only the first optimized reference $u_{0|k}$ is applied to the controller and the problem is re-solved at the next control step using the new measured state.
%----------------------------------------%
\section{Experiments}
To evaluate the Adaptive DA-KCF-MPC framework, we conduct experiments on a physical nano-quadrotor Crazyflie under nominal and aerodynamically disturbed conditions. 

\subsection{Experimental Setup and Implementation}
Hardware experiments are performed in an OptiTrack motion-capture arena using a Crazyflie 2.1 Brushless quadrotor. Pose measurements are streamed at approximately $100$~Hz, while DA-KCF-MPC runs off-board at $50$~Hz and sends optimized position and yaw references through Crazyswarm2~\cite{preiss2017crazyswarm} to the stock onboard cascaded PID controller\footnote{Experiments run on an off-board computer with an NVIDIA RTX 4080 Laptop GPU and Intel 14900KF CPU. MPC uses double-precision JAX-compiled L-BFGS-B.}. Aerodynamic disturbances are generated using the Real Weather Wind Tunnel fan array (see Fig.~\ref{fig:overview}).
%to an off-board computer equipped with an NVIDIA RTX 4080 Laptop GPU and an Intel 14900KF CPU, running ROS~2 and Crazyswarm2~\cite{preiss2017crazyswarm}. The KCF state combines motion-capture position and attitude with the Crazyflie's onboard-estimated velocities. The proposed controller operates off-board at $50$~Hz. DA-KCF-MPC optimizes position and yaw references transmitted through Crazyswarm2 to the stock onboard cascaded PID controller, which retains responsibility for low-level stabilization and motor actuation. Aerodynamic disturbances are generated using the Real Weather Wind Tunnel fan array.

We evaluate the proposed method under three hardware test scenarios. In all experiments, the fan array generates airflow along the inertial $x$-axis. First, uniform-flow trajectory-tracking experiments compare PID, Nominal KCF-MPC and Adaptive DA-KCF-MPC while the vehicle tracks the same figure-eight trajectory in the $x$--$z$ plane at constant yaw under $0$, $3.75$, and $5.00$~m/s flow. Second, a sustained-hover experiment under $5.00$~m/s flow evaluates online adaptation during closed-loop operation, including residual learning, candidate validation, and model deployment while the vehicle maintains a fixed hover reference. Third, a spatially varying wake experiment places a building-like obstacle between the fan array and the vehicle under a $2.50$~m/s free-stream condition. The obstacle modifies the incoming $x$-directed flow, and the vehicle follows a $\pm0.5$~m lateral trajectory along the $y$~axis through the resulting wake to evaluate adaptation as the disturbance varies with position.

For the trajectory-tracking experiments, the nominal and adaptive controllers form a controlled ablation: they use identical sensing, calibrated nominal dynamics, MPC objective, constraints, reference commands, and differ only in the prediction model used in Eq.~\eqref{eq:kcf_mpc}. This isolates the effect of online disturbance adaptation on model prediction and closed-loop tracking. Tracking performance is quantified using the three-dimensional position error $e_p(k):=\|p_k-p_k^d\|_2$ and trajectory position RMSE. Aggregate results are reported as mean $\pm$ one sample standard deviation across repeated trials, with representative trajectories shown separately. 
%Implementation details required for reproducibility, including network architectures, online-update settings, MPC execution and state reconstruction are provided in Appendix~\ref{app:implementation}

\begin{figure*}[!t]
    \centering
    \includegraphics[width=0.90 \textwidth]{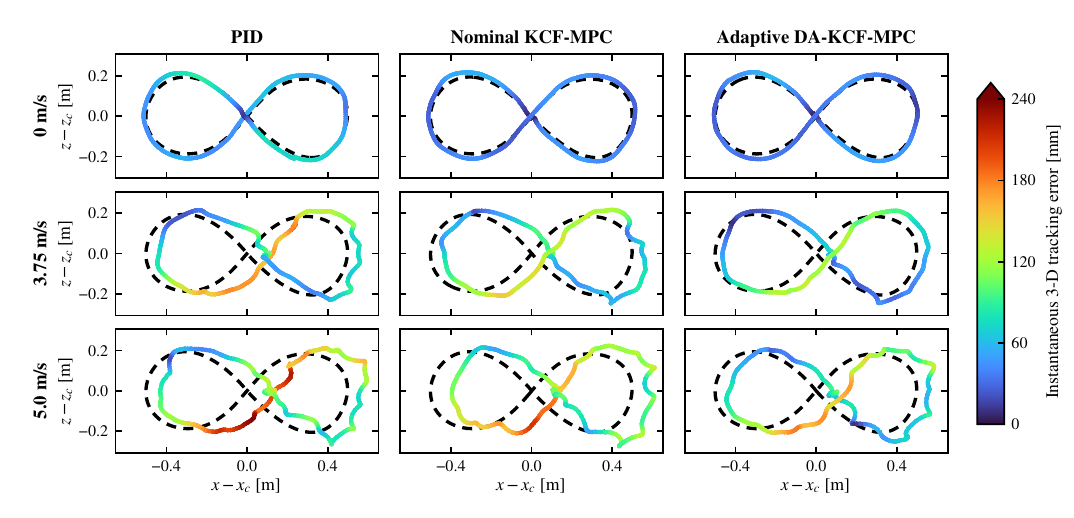}
    \vspace{-4mm}
    \captionsetup{font=footnotesize}
    \caption{Experimental figure-eight trajectories under increasing aerodynamic disturbance. Columns correspond to PID, Nominal KCF-MPC, and Adaptive DA-KCF-MPC, while rows correspond to fan-array flow speeds of $0$~m/s, $3.75$~m/s, and $5.0$~m/s. Trajectories are shown in the $x$--$z$ projection relative to the center $(x_c,z_c)$ of the desired trajectory. The dashed black curve denotes the desired path, while the measured trajectory is colored by the instantaneous three-dimensional position-tracking error $e_p(k)=1000\|p_k-p_k^d\|_2$ [mm]. Each panel shows a representative experimental trial. Aggregate performance over repeated trials is reported in Table~\ref{tab:tracking_rmse}.}
    \label{fig:tracking_grid}
    \vskip -0.2 true in
\end{figure*}
%--------------------------------------------%
\subsection{Results and Discussion}
\begin{table}[t]
\vspace{2mm}
\centering
\setlength{\tabcolsep}{10pt}
\captionsetup{font=footnotesize}
\caption{Three-dimensional position-tracking RMSE [cm] under increasing wind speed [m/s]. Nominal and Adaptive denote Nominal KCF-MPC and Adaptive DA-KCF-MPC, respectively. Values report mean $\pm$ one sample standard deviation across $4$ repeated trials.}
\label{tab:tracking_rmse}
\begin{tabular}{@{}cccc@{}}
\toprule
\textbf{Wind} &
\textbf{PID} &
\textbf{Nominal} &
\textbf{Adaptive} \\
\midrule
$0$
& $5.70 \pm 0.19$
& $3.66 \pm 0.18$
& $\mathbf{3.59 \pm 0.05}$ \\
$3.75$
& $10.72 \pm 0.08$
& $9.39 \pm 0.05$
& $\mathbf{7.70 \pm 0.33}$ \\
$5.00$
& $13.15 \pm 2.03$
& $12.63 \pm 0.27$
& $\mathbf{9.95 \pm 0.41}$ \\
\bottomrule
\end{tabular}
\end{table}
%-----------------------------------------------%
\begin{figure}[ht]
    \centering
    \includegraphics[width=0.95 \columnwidth]{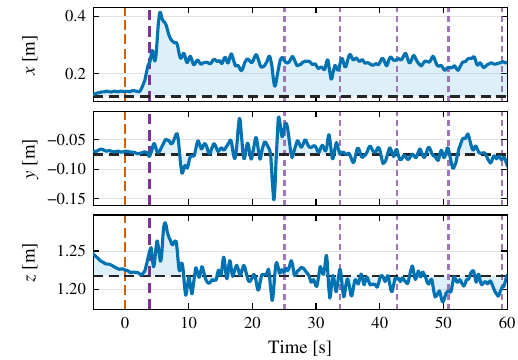}
    \captionsetup{font=footnotesize}
    \vskip -0.05 true in
    \caption{Online adaptation during sustained $5.0$~m/s wind hover. The orange dashed line marks disturbance onset, the dark-purple line the first adaptation trigger, and the light-purple lines subsequent candidate-update cycles. Dashed black lines denote the commanded hover position.}
    \label{fig:adaptive_hover}
    \vskip -0.2 true in
\end{figure}
%-----------------------------------------------%
We first evaluate whether online disturbance adaptation improves closed-loop trajectory tracking as model mismatch increases. Fig.~\ref{fig:tracking_grid} compares %representative
figure-eight paths for PID, Nominal KCF-MPC and Adaptive DA-KCF-MPC under increasing uniform flow, while Table~\ref{tab:tracking_rmse} reports aggregate performance across repeated trials. Under no wind, Nominal KCF-MPC and Adaptive DA-KCF-MPC achieve comparable tracking performance, with mean three-dimensional position RMSE of $3.66$ and $3.59$~cm, respectively. The $1.9\%$ difference indicates that online adaptation induces minimal change when the nominal predictor is already adequate. As the disturbance increases, the adaptive advantage becomes substantially larger. Relative to Nominal KCF-MPC, Adaptive DA-KCF-MPC reduces position RMSE by $18.0\%$ at $3.75$~m/s and by $21.2\%$ at $5.00$~m/s. At the same time, the advantage of the frozen nominal controller over PID decreases from $35.8\%$ under no wind to only $4.0\%$ at $5.00$~m/s, indicating that increasing disturbance erodes the benefit of the calibrated nominal model. Adaptive DA-KCF-MPC retains substantially lower tracking error under disturbance, with RMSE $28.2\%$ and $24.3\%$ below PID at $3.75$ and $5.00$~m/s, respectively. These results show that the proposed adaptation has little effect when nominal dynamics remain representative, but increasingly improves tracking as disturbance-induced model mismatch grows.

The sustained-hover experiment examines whether online adaptation can occur during closed-loop operation while the vehicle remains regulated under a persistent disturbance. Under $5.00$~m/s flow, wind onset produces a pronounced $x$~position deviation, followed by a persistent offset (Fig.~\ref{fig:adaptive_hover}). Despite repeated online training, the position error does not exhibit progressive growth, and the vehicle continues to regulate all three position coordinates throughout the experiment. The persistent $x$-offset therefore reflects a sustained disturbance-induced tracking bias rather than loss of closed-loop control. This result demonstrates that residual learning, validation, and model updates can be carried out online while the vehicle remains actively controlled.

\begin{figure}[t]
    \centering
    \includegraphics[width=0.95 \columnwidth]{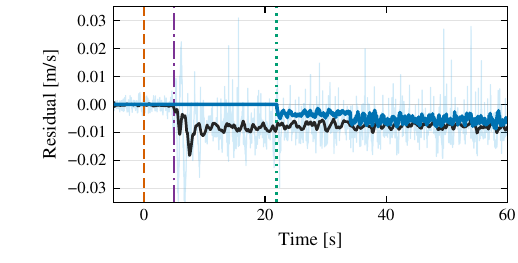}
    \captionsetup{font=footnotesize}
    \vskip -0.1 true in
    \caption{Online residual adaptation during sustained $5.0$~m/s wind hover. The light-blue trace shows the $x$-component of the observed one-step velocity prediction mismatch $\Delta v_{k,x}$, the black trace its causal $1$~s trailing mean, and the dark-blue trace the learned $x$-component of the ensemble correction $\Gamma_{k,x}$. Vertical lines indicate wind onset (orange), the first online training attempt (purple), and deployment of the first validated residual model (green). Time is referenced to wind onset.}
    \label{fig:online_residual}
    \vskip -0.15 true in
\end{figure}

Fig.~\ref{fig:online_residual} provides a model-level view of the same adaptation process. Here, $\Delta v_{k,x}$ denotes the $x$~component of the observed one-step velocity prediction mismatch, computed as the measured next-step velocity minus the corresponding prediction from the frozen nominal KCF in Eq.~\eqref{eq:calibrated_kcf}. Following disturbance onset, $\Delta v_{k,x}$ develops a persistent negative component together with substantial sample-to-sample variation. Prior to deployment, the learned correction $\Gamma_{k,x}$ remains zero. Once the first validated residual model is deployed, $\Gamma_{k,x}$ becomes nonzero with a sign consistent with the persistent prediction mismatch and remains bounded thereafter. The learned correction does not reproduce the observed residual's high-frequency fluctuations, indicating that the online learner extracts a repeatable disturbance-induced correction rather than tracking %instantaneous 
prediction noise.

\begin{figure}[t]
    \centering
    \includegraphics[width=\columnwidth, trim=0 38 0 0, clip]{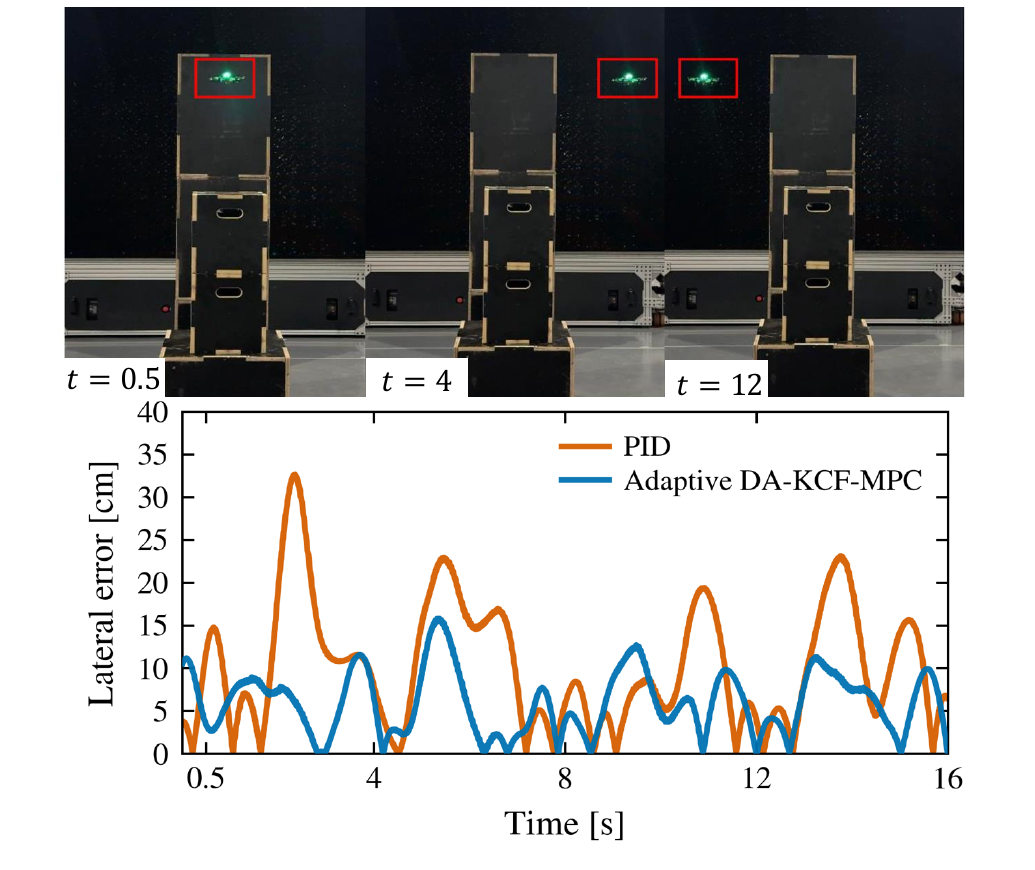}
    \captionsetup{font=footnotesize}
    \caption{Adaptive DA-KCF-MPC flight through a spatially varying wake generated by an obstacle representative of an urban building under a $2.5$~m/s
    free-stream condition. Top: representative snapshots at
    $t=\{0.5,4,12\}$~s as the vehicle follows the $\pm0.5$~m, $16$~s lateral trajectory along the $y$-axis through different regions of the wake. Bottom: lateral tracking error for representative PID and Adaptive DA-KCF-MPC trials. Aggregate tracking performance is evaluated over 4 repeated trials per controller.}
    \label{fig:smoothY_wake_photos}
    \vskip -0.1 true in
\end{figure}

Finally, we evaluate whether the proposed adaptation extends from approximately uniform flow to a spatially varying disturbance field. A building-like obstacle is placed between the fan array and the vehicle under a $2.50$~m/s free-stream condition, causing the incoming $x$-directed airflow to form a nonuniform downstream wake. The vehicle follows a smooth $\pm0.5$~m lateral trajectory along the $y$-axis over $16$~s, thereby encountering position-dependent disturbance conditions. This setting is more challenging than uniform flow because the disturbance experienced by the vehicle varies with position and therefore cannot, in general, be represented by a single constant correction over the trajectory. As progressively more of the wake is encountered, successive residual models are learned and validated online. After exposure to the full lateral range, including both positive and negative $y$-directions, the validated adaptive model reduces the wake-specific aggregate multi-horizon velocity prediction RMSE by $12.9\%$ relative to the frozen nominal predictor. Across four repeated hardware trials, Adaptive DA-KCF-MPC reduces $y$-axis tracking RMSE from $11.55$~cm with PID to $7.94$~cm, a $31.3\%$ reduction. Three-dimensional position RMSE similarly decreases from $14.38$ to $11.03$~cm, a $23.3\%$ reduction (Fig.~\ref{fig:smoothY_wake_photos}). Together, the prediction and tracking improvements indicate that the learned correction captures repeatable, position-dependent effects of the wake rather than merely compensating for a
constant wind bias.
%---------------------------------------------%
%--------------------------------------------%

To determine whether the observed tracking improvements are accompanied by improved model prediction, Tables~\ref{tab:position_ablation} and~\ref{tab:velocity_ablation} compare the simulation-identified KCF, the hardware-calibrated KCF and the adaptive DA-KCF using identical measured initial states and recorded reference sequences. At the $h_p=25$~step prediction horizon used by the MPC, hardware data calibration through LoRA reduces position and velocity RMSE by $26.3\%$ and $61.5\%$, respectively, compared to the identified model from purely simulation data. Online adaptation further reduces these errors by $7.6\%$ and $21.0\%$ relative to the LoRA-calibrated predictor. Thus, the final adaptive model achieves the lowest position and velocity prediction error at the full MPC horizon. The shorter-horizon results are less uniform. For position prediction, calibration and adaptation incur small degradations at $h_p=\{1,5,10\}$ relative to the simulation-identified model, but substantially reduce error at the full $25$-step MPC horizon. Velocity prediction improves after adaptation at $h_p=1$ and $5$, degrades modestly relative to the calibrated model at
$h_p=10$, and improves again at $h_p=25$. Thus, the Adaptive DA-KCF model does not uniformly outperform the nominal KCF at every intermediate step; its principal benefit emerges over the full prediction horizon used for control.
%-------------------------------------------%
\begin{table}[t]
\vspace{2mm}
\centering
\small
\setlength{\tabcolsep}{7pt}
\captionsetup{font=footnotesize}
\caption{Multi-step position prediction RMSE [mm] on held-out disturbed hardware sequences using identical measured initial states and recorded reference commands.}
\label{tab:position_ablation}
\begin{tabular}{lccc}
\toprule
\textbf{Horizon $h_p$} &
\textbf{Sim. KCF} &
\textbf{Cal. KCF} &
\textbf{Adap. DA-KCF} \\
\midrule
$1$  & 2.48   & 2.83  & 2.83 \\
$5$  & 10.58  & 12.31 & 12.43 \\
$10$ & 22.59  & 24.28 & 24.82 \\
$25$ & 120.58 & 88.83 & \textbf{82.09} \\
\bottomrule
\end{tabular}
\vskip -0.0 true in
\end{table}
\begin{table}[!t]
\centering
\small
\setlength{\tabcolsep}{7pt}
\captionsetup{font=footnotesize}
\caption{Multi-step velocity prediction RMSE [mm/s] on held-out disturbed hardware sequences using identical measured initial states and recorded reference commands.}
\label{tab:velocity_ablation}
\begin{tabular}{lccc}
\toprule
\textbf{Horizon $h_p$} &
\textbf{Sim. KCF} &
\textbf{Cal. KCF} &
\textbf{Adap. DA-KCF} \\
\midrule
$1$  & 9.42   & 6.89   & \textbf{4.97} \\
$5$  & 46.18  & 27.04  & \textbf{21.46} \\
$10$ & 102.57 & \textbf{40.08} & 45.64 \\
$25$ & 486.23 & 186.96 & \textbf{147.79} \\
\bottomrule
\end{tabular}
\vskip -0.2 true in
\end{table}
The resulting Adaptive DA-KCF-MPC controller also satisfies the $50$~Hz real-time requirement. Across $7{,}458$ MPC updates during the $5.0$~m/s hover experiment, the median, $95$th-percentile, and maximum solve times were $10.84$, $11.52$, and $14.44$~ms, respectively, all below the $20$~ms control period. Online adaptation executes asynchronously from MPC and therefore does not block the control loop. The median end-to-end adaptation duration was $6.74$~s, with median residual-optimization and validation times of $4.93$ and $1.77$~s, respectively; the maximum observed total adaptation duration was $8.43$~s.

Collectively, these hardware results demonstrate that the calibrated KCF provides accurate nominal reference-to-state prediction, while online adaptation becomes increasingly beneficial as disturbance-induced model mismatch grows. The adaptive correction improves prediction over the control-relevant horizon, translates this improvement into lower trajectory-tracking error under both uniform and spatially varying disturbances and operates without violating the deployed real-time control requirement. Representative hardware trials and online adaptation behavior are provided in the accompanying supplementary video.
%----------------------------------------%
\section{Conclusion and Future Work} 
\label{sec:conc}
This work presented the first real-time demonstration of an online disturbance-aware Koopman Control Family (DA-KCF). The proposed framework learns a reference-dependent nominal KCF in simulation, calibrates it using hardware data and learns a low-dimensional disturbance correction online while keeping the nominal model fixed. Hardware experiments demonstrate improved multi-step prediction and trajectory tracking while satisfying the $50$~Hz real-time control requirement. Future work will investigate richer matrix-valued disturbance operators beyond the structured velocity correction, multidirectional and faster-varying disturbances, and higher-performance onboard computation for more expressive online adaptation.   
%----------------------------------------%
\bibliographystyle{IEEEtran}
\bibliography{Bib/refs}
%\clearpage
%\input{appendix}
\end{document}